\pdfoutput=1
\documentclass[sigconf, screen]{acmart}

\copyrightyear{2019}
\acmYear{2019} 
\setcopyright{iw3c2w3}
\acmConference[WWW '19]{Proceedings of the 2019 World Wide Web Conference}{May 13--17, 2019}{San Francisco, CA, USA}
\acmBooktitle{Proceedings of the 2019 World Wide Web Conference (WWW '19), May 13--17, 2019, San Francisco, CA, USA}
\acmPrice{}
\acmDOI{10.1145/3308558.3313650}
\acmISBN{978-1-4503-6674-8/19/05}

\usepackage{booktabs} % For formal tables
\usepackage{arydshln}
\usepackage{subcaption,verbatim,engord,enumitem,multirow,bm,caption}
\usepackage{algorithm,algorithmic}
\usepackage{amsmath,dsfont}
\usepackage{amssymb}
\usepackage{graphicx}
\usepackage{amscd}
\usepackage{mathtools}
\usepackage{balance}

\begin{document}
% [CAMERA-READY] Change the title to be consistent to the one in submission system.
\title{Towards Neural Mixture Recommender for Long Range Dependent User Sequences}

\author{Jiaxi Tang$^{1*+}$, Francois Belletti$^{2+}$, Sagar Jain$^2$, Minmin Chen$^2$, Alex Beutel$^2$, Can Xu$^2$, Ed H. Chi$^2$ }\thanks{$^*$ Work done while interning at Google}\thanks{$^+$The two authors contributed equally to this work}
\affiliation{ $^1$School of Computing Science, Simon Fraser University, Canada \ \  $^2$Google}
\affiliation{$^1$jiaxit@sfu.ca \ \ $^2$ \{belletti, sagarj, minminc, alexbeutel, canxu, edchi\}@google.com}

\fancyhead{} 

\begin{abstract}
Understanding temporal dynamics has proved to be highly valuable for accurate recommendation. 
Sequential recommenders have been successful in modeling the dynamics of users and items over time. 
However, while different model architectures excel at capturing various temporal ranges or dynamics, distinct application contexts require adapting to diverse behaviors.
%On recommendation platforms with a large and diverse catalog of items, the key user experience is to be able to discover which items are available that suit their needs and tastes. Providing accurate recommendations based on sequences of previously observed user/item interactions is therefore a crucial task that is challenged by the length of such sequences and the fact that user behaviors are often influenced by their long-term memory. 

%In this paper, we highlight the challenge of capturing latent interests in long sequences of user history from a production dataset comprising of millions of sequences up to 300 element long. 
In this paper we examine how to build a model that can make use of different temporal ranges and dynamics depending on the request context.
% We adopt several novel approaches to address this problem.
We begin with the analysis of an anonymized Youtube dataset comprising millions of user sequences. % up to 300 element long.
We quantify the degree of long-range dependence in these sequences and demonstrate that both short-term and long-term dependent behavioral patterns co-exist.
We then propose a neural Multi-temporal-range Mixture Model~(\emph{M3}) as a tailored solution to deal with both short-term and long-term dependencies.  
Our approach employs a mixture of models, each with a different temporal range. %hese models are combined by a learned gating mechanism capable of using different combinations of models to provide contextual recommendations for our home page as well as item-specific pages. 
These models are combined by a learned gating mechanism capable of exerting different model combinations given different contextual information.
In empirical evaluations on a public dataset and our own anonymized YouTube dataset, \emph{M3} consistently outperforms state-of-the-art sequential recommendation methods.
\end{abstract}




\maketitle

\section{introduction}
\label{sec:intro}
Across the web and mobile applications, recommender systems are relied upon to surface the right items to users at the right time.
Some of their success can be attributed to advances in modeling as well as the ingenuity of applied researchers in adopting and inventing new techniques to solve this important problem~\cite{sarwar2001item,koren2009matrix,rendle2009bpr,he2017neural}. 
Fundamentally, recommenders match users in a particular context with the best personalized items that they will engage with~\cite{linden2003amazon,gomez2016netflix}.
In order to do this effectively, recommenders need to understand the users, typically based on their previous actions, and to understand items, most often based on the users who previously interacted with them. %understand the context of the request. 
This presents a fundamental challenge: users' preferences and items' perception are continuously changing over time, and the recommender system needs to understand these dynamics.

A significant amount of research has recognized forms of this problem.
Sequence information has been generally shown to improve recommender performance~\cite{HeMcA16b,rendle2010factorizing}.
%have demonstrated that sequence information improves a recommender's performance.
\citet{koren2009collaborative} identified multiple user and item dynamics in the Netflix Prize competition, and incorporated these dynamics as biases in a collaborative filtering model.
\cite{wu2017rnn, beutel2018latent} demonstrated that Recurrent Neural Networks (RNNs) could learn many of these patterns, and likewise \cite{hidasi2015session} demonstrated that RNNs can learn patterns in individual sessions. 
Despite these successes, RNNs are known to have difficulties learning long-range dependent temporal patterns~\cite{belletti2018factorized}.

We observe and study an open challenge for such sequential recommender systems:
\emph{
while different applications and contexts require different temporal ranges and patterns, model architectures are typically designed to capture a particular temporal dynamic.
}
For example, when a user comes to the Amazon home page they may be looking for something new to buy or watch, but on an item specific page they may be looking for other items that are closely related to recently browsed items.
\emph{How can we design a model that works, simultaneously, across all of these contexts and temporal ranges?}

\textbf{Contributions:} 
We address the issue of providing a single model adapted to the diversity of contexts and scales of temporal dependencies in sequential recommendations through data analysis and the design of
a Multi-temporal-range Mixture Model, or \emph{M3} for short.
We make the following contributions to this problem:
\begin{itemize}
\item
    \textbf{Data-driven design:} We demonstrate that in real world recommendation tasks there are significant long-range temporal dependencies in user sequence data, and that previous approaches are limited in their ability to capture those dynamics. M3's design is informed by this quantitative analysis.
\item
    \textbf{Multi-range Model:} We offer a single model, M3, which is a mixture model consisting of three sub-models (each with a distinct manually designed architecture) that specialize in capturing different ranges of temporal dependencies. M3 can learn how to dynamically choose to focus on different temporal dynamics and ranges depending on the application context. 
\item 
    \textbf{Empirical Benefits and Interpretability:}
    We show on both public academic and private data that our approach provides significantly better recommendations. Further, using its interpretable design, we analyze how M3 dynamically switches between patterns present at different temporal ranges for different contexts, thus showing the value in enabling context-specific multi-range modeling.
    Our private dataset consists in anonymized user sequences from YouTube. To the best of our knowledge this paper is the first to focus on sequential patterns in such a setting.
\end{itemize}

% % %\vspace{-0.2cm}

\section{Related Work}\label{sec:related}
Before we describe our sequential recommendation problem and provide the quantitative insights orienting the design of a novel sequential neural model based on a mixture of models, we briefly introduce the reader to some key pre-existing related work.

Matrix factorization~\cite{koren2009matrix} is among the most popular techniques used in classic recommender research, in which a similarity score for each user-item pair is learned by building latent user and item representations to recover historical user-item interactions. 
The predicted similarity score is then used to indicate the \emph{relatedness} and find the most relevant items to recommend to a user. %between user $u$ and item $v$ which later on help determine which items are most relevant to the user and should be recommended.
% Context and time are typically not modeled in these settings.
Followup work on introducing auxiliary sources of information beyond user-item interactions have been proven successful~\cite{paul2016deep}, especially for cold-start problems. \citet{pazzani2007content} use item content (\emph{e.g.,} product image, video's visual/audio content, etc) to provide a better item representation. 

\textbf{Neural Recommender Systems.} Deep neural networks have gained tremendous success in the fields of Computer Vision~\cite{karpathy2014large, krizhevsky2012imagenet} and Natural Language Processing~\cite{bahdanau2014neural, mikolov2010recurrent}. In recommender research, we have witnessed growing interest of using deep neural networks to model complex contextual interactions between user and items, which surpass classic factorization-based methods~\cite{koren2009matrix,rendle2010factorization}. 
Auto-encoders\cite{sedhain2015autorec, wu2016collaborative, liang2018variational} constitute an early example of success for a framework based on neural networks to better infer un-observed user/item affinities in a recommendation problem. % in a recommendation problem as they have been able to suitably solve the task of inferring un-observed user/item affinities. 
\citet{he2017neural} also proved that traditional Collaborative Filtering methods can be effectively generalized by a deep neural network. Besides, 

For the specific problem of sequential recommendation using neural networks, RNNs~\cite{hidasi2015session, wu2017rnn} have become a common choice. 
Other methods based on Convolutional Neural Networks (CNNs)~\cite{tang2018caser,yuan2019simple}, Attention Models~\cite{zhou2018deep} have also been explored. 
While most of existing methods developed for sequential recommendations perform well~\cite{HeMcA16b, rendle2010factorizing, tang2018caser, hidasi2015session, smirnova2017context}, they still have some limitations when dealing with long user sequences found in production recommender systems. 
As we shall discuss in Section~\ref{sec:pre}, such approaches do not scale well to very long sequences. 

\textbf{Mixture of Models.}
Despite being simpler and more elegant, monolithic models are in general less effective than mixtures of models to take advantage of different model capacities and architectural biases. 
\citet{gehring2017convolutional} used an RNN in combination with an attention model for neural machine translation which provided a substantial performance gain.
\citet{pinheiro2014recurrent} proposed to combine a CNN with an RNN for scene labeling.  
In the field of sequential recommendation, an earlier work on mixing of a Latent Factor Model~(LFM) and a Factorized Markov Chain~(FMC) has been shown to offer superior performance than each individual one~\cite{rendle2010factorizing}. A similar trend was observed in~\cite{HeMcA16b,zhou2018dien}.
While sharing similar spirit to these aforementioned methods, we designed our mixture of models with the goal to model varying ranges of dependence in long user sequences found in real production systems. 
Unlike model ensembles~\cite{dietterich2000ensemble, zhou2002ensembling,zhu2018brand} that learn individual models separately prior to ensembling them, a mixture of models learns individual models as well as combination logic simultaneously.

\section{Quantifying Long Range Temporal Dependencies in User Sequences}\label{sec:pre}

We first present some findings on our anonymized proprietary dataset which uncover properties of behavioral patterns as observed in extremely-long user-item interaction sequences. We then pinpoint some limitations of existing methods which motivate us to design a better adapted solution.

\begin{figure}  %figure divergent

    \includegraphics[width=0.47\textwidth]{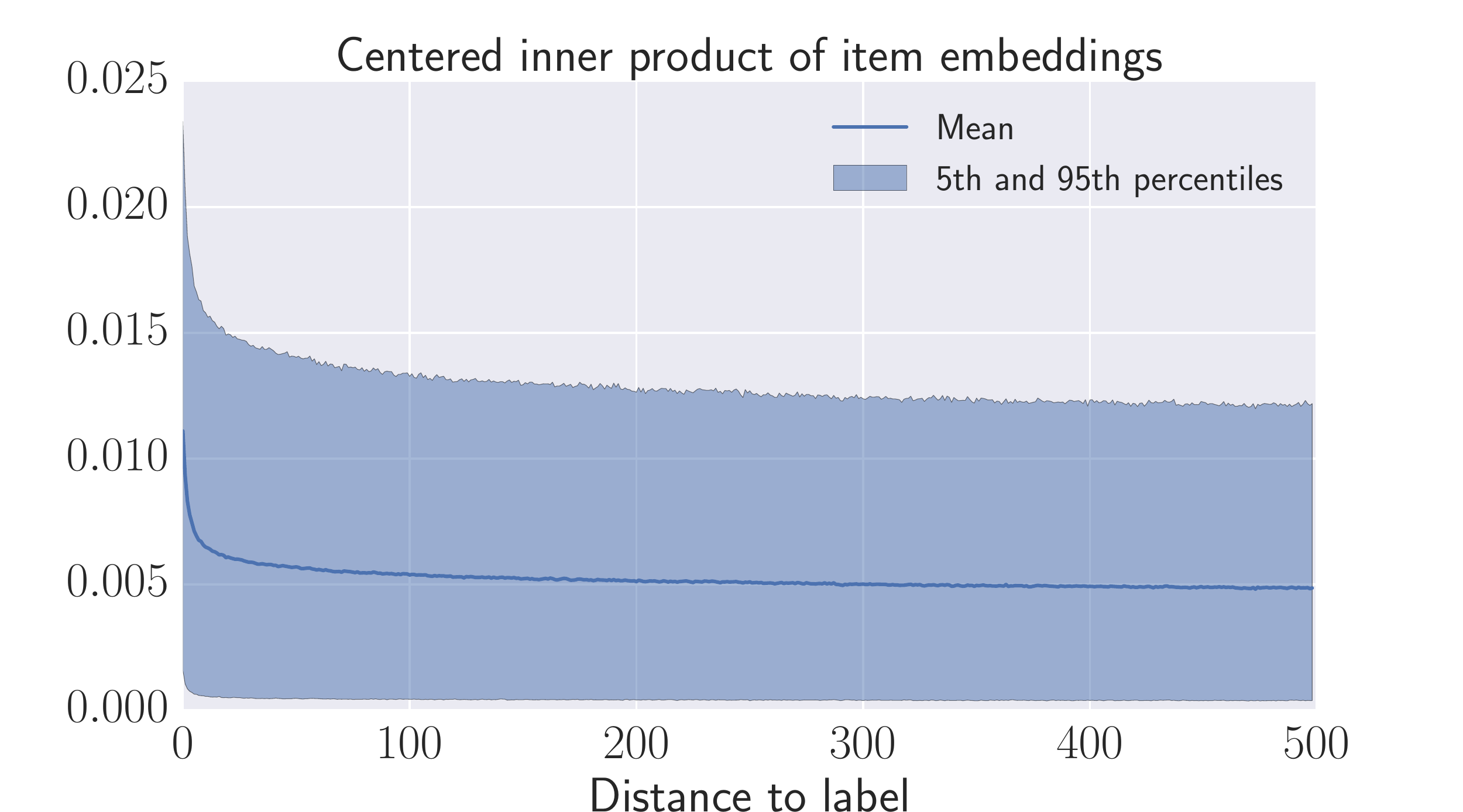}
    % \caption{$L \in [1,499]$, 100K samples}
    \label{fig:cov_zoomout2}
    \vspace{-0.25in}
    \caption{Trace of covariance (i.e. centered inner product similarity) of item embeddings between the last item in user sequence and the item located $L$ steps before (100K samples).}\label{fig:cov_zoomout}

\end{figure}

\textbf{Sequential Recommendation Problem:}
We consider a sequential recommendation problem~\cite{rendle2010factorizing, HeMcA16b, tang2018caser, hidasi2015session,smirnova2017context} defined as follows: assume we have a set of users $u \in \mathcal{U}$, a set of items $v \in \mathcal{V}$, and for each user we have access to a sequence of user historical events $\mathcal{E}^u=(e^{u}_{1},e^{u}_{2},\cdots)$ ordered by time. %where $\mathcal{E}^u$ is the sequence of user events ordered by time, %$N^{(u)}=|\mathcal{E}^u|$ is the length of a event sequence and
Each $e^u_\tau$ records the item consumed at time $\tau$ as well as context information of the interaction. % is the event that occurred at step $\tau$.
Given the historical interactions, our goal is to recommend to each user a subset of items in order to maximize a performance metric such as user satisfaction.

\subsection{Hidden Values in Long User Sequences}
We now describe how we developed a better understanding of long user sequences in our proprietary dataset through quantitative data exploration.
To quantify how past events can influence a user's current behavior in our internal dataset, \emph{i.e.} measure the range of temporal dependency within a sequence of events, one can examine the covariance matrix of two events $L$-step apart~\cite{belletti2017random,pipiras2017long}, 
where step denotes the relative order of events within sequence.
In particular, we look at the trace of the covariance matrix as a measurement of dependency: %we measure the range of temporal dependency within a sequence of events with the same similarity function as the one we employ to characterize user/item affinity in our sequential recommender.
% Namely, we compute:
\begin{equation*}\label{equ:cov}
\text{Dep}_L
%=
%E\left[ Q_{e_N}^T Q_{e_{N-L}} \right] - E\left[ Q_{e_N} \right]^T E \left[ Q_{e_{N-L}} \right]
=
\text{tr} \left(\text{Cov}\left(Q_{e_N}, Q_{e_{N-L}}\right)\right)
\end{equation*}
where ${e_N}$ is the item in last event in a logged user/item interaction sequence and ${e_{N-L}}$ is the item corresponding to the interaction that occurred $L$ time steps before the last event.  We focus on the trace of the covariance matrix %between $Q_{e_N}$ and $Q_{e_{N-L}}$ 
as it equals the sum of the eigenvalues of the covariance matrix and its rate of decay is therefore informative of the rate of decay of these eigenvalues as a whole.

We utilize the embeddings $Q$ that have been learned by a pre-existing model---in our case an RNN-based sequential recommender which we describe later as one of M3's sub-models. $\text{Dep}_L$ here measures the similarity between the current event and the event $L$ steps back from it. 
To estimate $\text{Dep}_L$ for a particular value of $L$ we employ a classic empirical averaging across user sequences in our dataset. % of the observed similarity scores.
From Figure~\ref{fig:cov_zoomout}, we can extract multiple findings: 
\begin{itemize}
\item The dependency between two events decreases as the time separating their consumption grows. This suggests that recent events bear most of the influence of past user behavior on a user's future behavior.
\item The dependency slowly approaches zero even as the temporal distance becomes very large (\emph{i.e.} $L>100$).
The clear hyperbolic-decay of the level of temporal dependencies indicates the presence of long-range-dependent patterns existing in user sequences~\cite{pipiras2017long}.
In other words, a user's past interactions, though far from the current time step, still cumulatively influence their current behavior significantly.
\end{itemize}
These findings suggest that users do have long-term preferences and better capturing such long-range-dependent pattern could help predicting their future interests.
In further work, we plan to use off-policy correction methods such as~\cite{gilotte2018offline,chen2019top} to remove presentation bias when estimating correlations.

\subsection{Limitations of Existing Sequential Models}
The previous section has demonstrated the informational value of long-range temporal patterns in user sequences. 
Unfortunately, it is still generally challenging for existing sequential predictive models to fully utilize information located far into the past.

\emph{Most prior models have difficulties when learning to account for sequential patterns involving long-range dependence}.
Existing sequential recommenders with factorized Markov chain methods~\cite{HeMcA16b} or CNNs~\cite{tang2018caser} arguably provide reliable sequential recommendation strategies. 
Unfortunately they are all limited by a short window of significant temporal dependence when leveraging sequential data to make a recommendation prediction.
RNNs~\cite{hidasi2015session, jannach2017recurrent, beutel2018latent} and their variants~\cite{quadrana2017personalizing, devooght2017long} are widely used in sequential recommendation.
RNN-based models, though effective for short user sequences~(\emph{e.g.} short event sequences within a session), are challenged by long-range dependent patterns in long user sequences. 
Because of the way they iterate over sequential items~\cite{mikolov2010recurrent} and their use of saturating non-linear functions such as $\tanh$ to propagate information through time, RNNs tend to have difficulties leveraging the information contained in states located far into the past due to gradient propagation issues~\cite{pascanu2013difficulty,belletti2018factorized}. 
Even recent architectures designed to facilitate gradient propagation such as Gated Recurrent Unit~\cite{cho2014learning} and Long-short Term Memory~\cite{hochreiter1997long, sutskever2014sequence} have also been shown to suffer from the same problem of not being able to provably account for long-range dependent patterns in sequences~\cite{belletti2018factorized}.

\emph{A second challenge in sequential recommendations is learning user latent factors $P_u$ explicitly from data}, which has been observed to create many difficulties~\cite{tang2018caser, HeMcA16b, chen2018sequential,rendle2010factorizing}. 
In the corresponding works, users' long-term preferences have been modeled through learning a set of latent factors $P_u$ for each user. 
However, learning $P_u$ explicitly is difficult in large-scale production systems. As the number of users is usually several magnitudes higher than the number of items, building such a large user vocabulary and storing the latent factors in a persistent manner is challenging.
Also, the long-tail users (\emph{a.k.a} cold users) and visitor users (\emph{i.e.} users who are not logged in) could have much worse recommendations than engaged users~\cite{beutel2017beyond}.

\subsection{Limitations of Single Monolithic Models} 
Figure~\ref{fig:cov_zoomout} clearly indicates that although the influence of past user events on future interactions follows a significant decaying trend, significant predictive power can still be carried by events located arbitrarily far in the past.
Very recent events~(\emph{i.e.} $1 \le L \le 10$) have large magnitude similarities with the current user behavior and this similarity depends strongly on the sequential order of related events. 
As the distance $L$ grows larger, the informative power of previously consumed items on future user behavior is affected by more uncertainty (\emph{e.g.} variance) and is less sensitive to relative sequential position. 
That is, the events from 100 steps ago and from 110 steps ago may have a generally similar influence on future user decisions regardless of their relative temporal location.
Therefore, \emph{for the kind of sequential signals we intend to leverage, in which different scales of temporal dependencies co-exist, it may be better to no longer consider a single model}. 
While simple monolithic models such as Deep Neural Network (DNN) with pooling and dropout~\cite{wu2017session, paul2016deep} are provably robust to noise, they are unfortunately not sensitive to sequential order (without substantial modifications).
On the other hand, RNNs~\cite{hidasi2015session, beutel2018latent} provide cutting-edge sequential modeling capabilities but they are heavily sensitive to noise in sequential patterns.
Therefore, it is natural to choose a mixture of diverse models which would then complement each other to provide better overall predictive power.

\section{Multi-temporal-range Mixture Model for Long User Sequences}
\label{sec:model}

Motivated by our earlier analyses, we now introduce a novel method aimed at addressing the shortcoming of pre-existing approaches for long user/item interaction sequences: Multi-temporal-range Mixture Model~(M3) and its two variants~(M3R/M3C).
For simplicity, we omit the superscripts related to users (\emph{i.e.} $e^u_\tau$ will now be denoted $e_\tau$) and use a single user sequence to describe the neural architecture we introduce.

\subsection{Overview}

\begin{figure}[ht!]  %figure divergent
\centering
\includegraphics[scale=0.55]{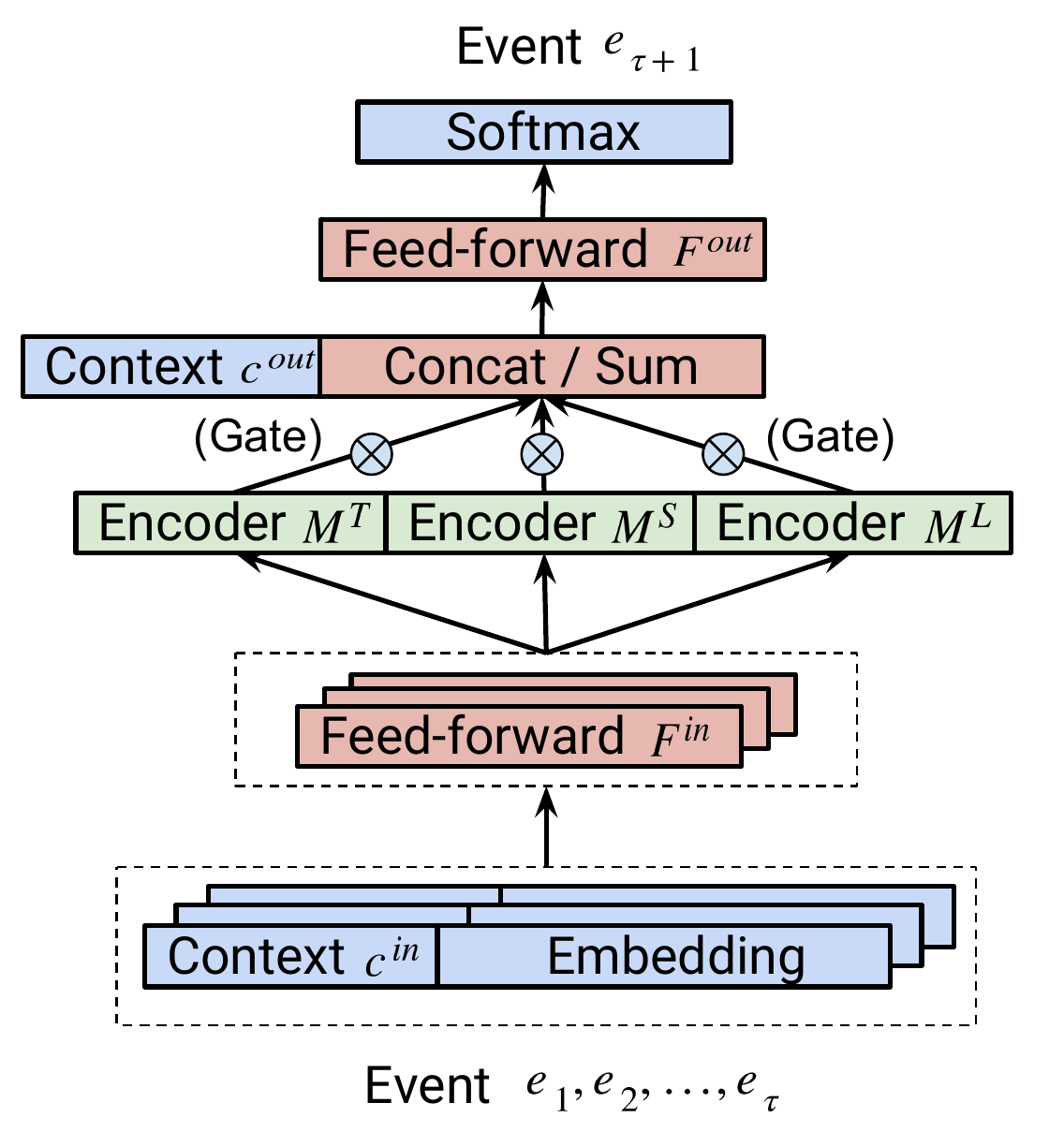}
\vspace{-.25cm}
\caption{ An overview of the proposed M3
model.}
\label{fig:network_overview}
\end{figure}

Figure~\ref{fig:network_overview} gives a general schematic depiction of M3. 
We will now introduce each part of the model separately in a bottom-up manner, starting from the inputs and progressively abstracting their representation which finally determines the model's output.
When predicting the behavior of a user in the next event $e_{\tau+1}$ of their logged sequence, we employ item embeddings and context features~(optional) from past events as inputs:
\begin{equation}\label{eqn:input}
x_{\tau} = [Q_{\tau} \oplus c^{\textrm{in}}_\tau],
\end{equation}
where $\oplus$ denotes the concatenation operator.
To map the raw context features and item embeddings to the same high-dimensional space for future use, a feed-forward layer $F^{\textrm{in}}$ is used:
\begin{equation}
    Z^{\textrm{in}}_{\tau} = \{z^{\textrm{in}}_i\}_{i=1\cdots \tau}
    \text{ where }
    z^{\textrm{in}}_{\tau} = F^{\textrm{in}}(x_{\tau})
\end{equation}
here $z^{\textrm{in}}_\tau \in \mathbb{R}^{1 \times d_{\textrm{in}}}$ represents the input processed at step $\tau$ and $Z^{\textrm{in}}_\tau \in \mathbb{R}^{\tau \times d_{\textrm{in}}}$ stands for the collection of all processed inputs before step $\tau$ (included).
Either the identity function or a ReLU~\cite{nair2010rectified} can be used to instantiate the feed-forward layer $F^{\textrm{in}}$.

In the previous section, we assessed the limitations of using a single model on long user sequence.
To circumvent the issues we highlighted, we employ in M3 three different sequence models~(encoders) in conjunction, namely $M^T$, $M^S$ and $M^L$, on top of the processed input $Z^{\textrm{in}}_{\tau}$.
We will later explain their individual architectures in details.
The general insight is that we want each of these sub-models to focus on different ranges of temporal dependencies in user sequences to provide a better representation~(\emph{i.e.,} embedding) of the sequence.
We want the sub-models to be architecturally diverse and address each other's shortcomings.
Hence
% \begin{equation*}
% SE_{\tau} = \{ SE^T_\tau, SE^S_\tau, SE^L_\tau \},
% \end{equation*}
% where 
\begin{equation}
SE^T_{\tau} = M^T(Z^{\textrm{in}}_\tau),~SE^S_{\tau} = M^S(Z^{\textrm{in}}_\tau),~ SE^L_{\tau} = M^L(Z^{\textrm{in}}_\tau),
\end{equation}
which yields three different representations, one produced by each of the three sequence encoders.
The three different sub-model encoders are expected to produce outputs---denoted by $d_{\textrm{enc}}$---of identical dimension.
By construction, each sequential encoder produces its own abstract representation of a given user's logged sequence, providing diverse latent semantics for the same input data.

Our approach builds upon the success of Mixture-of-Experts~(MOE) model~\cite{jacobs1991adaptive}. 
One key difference is that our `experts' are constructed to work with different ranges of temporal dependencies, instead of letting the cohort of `experts' specialize by learning from data.
As shown in \cite{shazeer2017outrageously}, heavy regularization is needed to learn different experts sharing the same architecture in order to induce specialization and prevent starvation when learning (only one expert performs well because it is the only one to learn which creates a self-reinforcing loop when learning with back-propagation).

Informed by the insights underlying the architecture of MOE models, we aggregate all sequence encoders' results by weighted-concatenate or weighted-sum, with weights $G_\tau$ computed by a small gating network.
In fact, we concatenate the outputs with
\begin{equation}
SE_{\tau} = (G^{T}_\tau \times SE^T_\tau) \oplus (G^{S}_\tau \times SE^S_\tau) \oplus (G^{L}_\tau \times SE^L_\tau),
\end{equation}
where $G_{\tau} \in \mathbb{R}^{3}$ corresponds to the outputs of our gating network. 
We can also aggregate outputs with a weighted-sum:
\begin{equation}
SE_{\tau} = (G^{T}_\tau \times SE^T_\tau) + (G^{S}_\tau \times SE^S_\tau) + (G^{L}_\tau \times SE^L_\tau).
\end{equation}
Note that there is no theoretical guarantee whether concatenation is better than summation or not. 
The choice of aggregation, as well as the choice of activation functions, is determined by observing a given model's performance from a validation set extracted from different datasets. 
Such a procedure is usual in machine learning and will help practitioners determine which variant of the model we propose is best suited to their particular application.

\begin{figure*}[bt]  %figure divergent

        \centering
        \begin{subfigure}[b]{0.25\textwidth}
                \includegraphics[width=\textwidth]{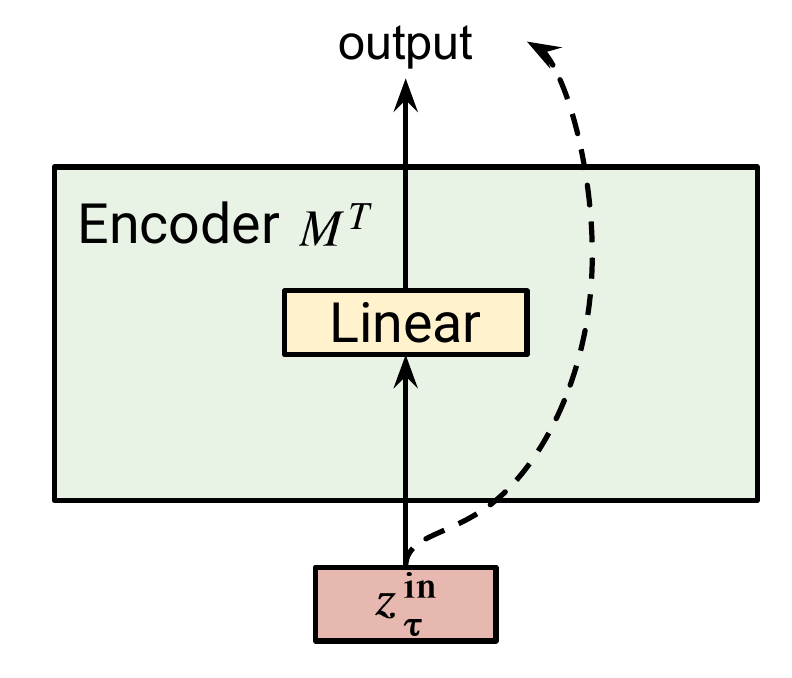}
                \caption{Tiny-range encoder}
                \label{fig:encoders_mt}
        \end{subfigure}
        ~
        \begin{subfigure}[b]{0.25\textwidth}
                \includegraphics[width=\textwidth]{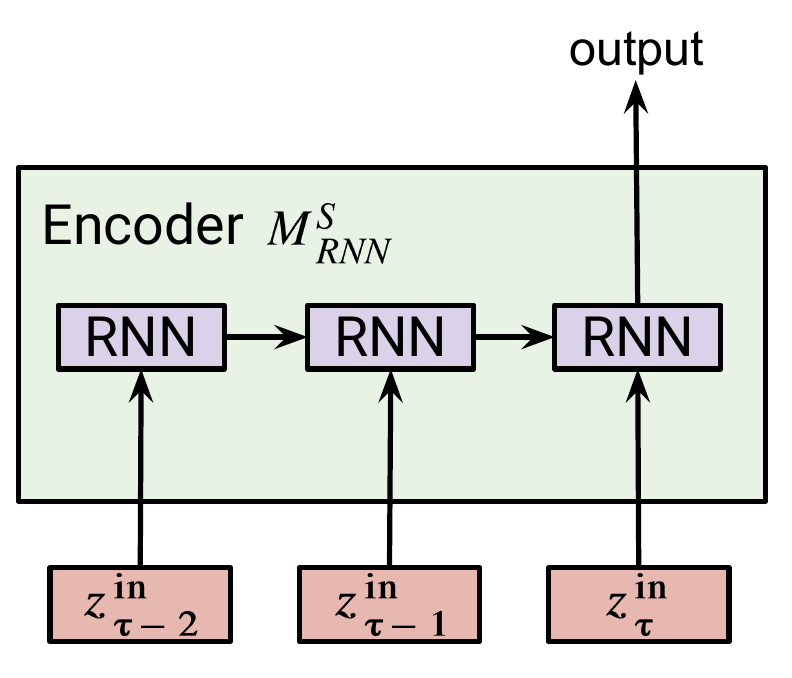}
                \caption{Short-range encoder~(RNN)}
                \label{fig:encoders_rnn}
        \end{subfigure}
        ~
        \begin{subfigure}[b]{0.25\textwidth}
                \includegraphics[width=\textwidth]{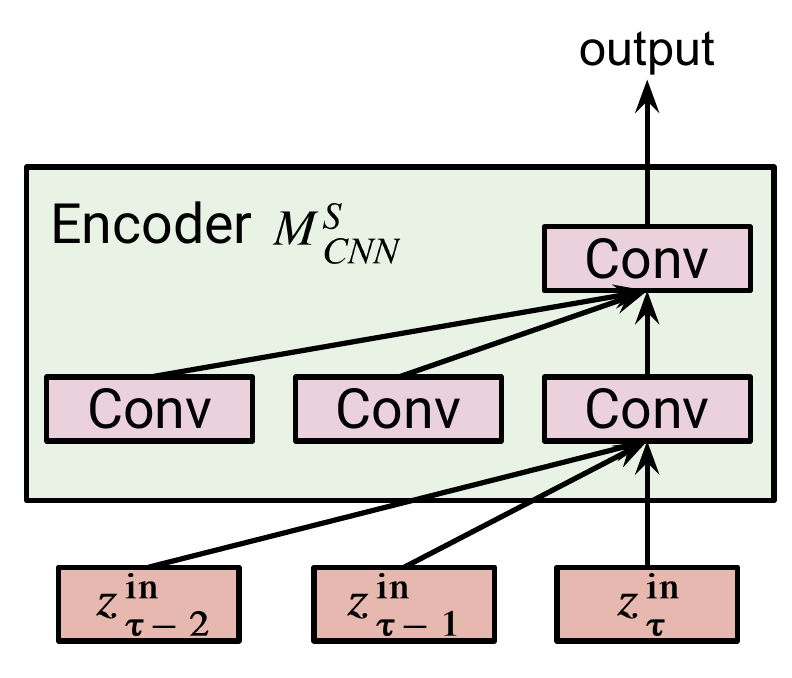}
                \caption{Short-range encoder~(CNN)}
                \label{fig:encoders_cnn}
        \end{subfigure}
        ~
        \hspace{0.02in}
        \begin{subfigure}[b]{0.25\textwidth}
                \includegraphics[width=\textwidth]{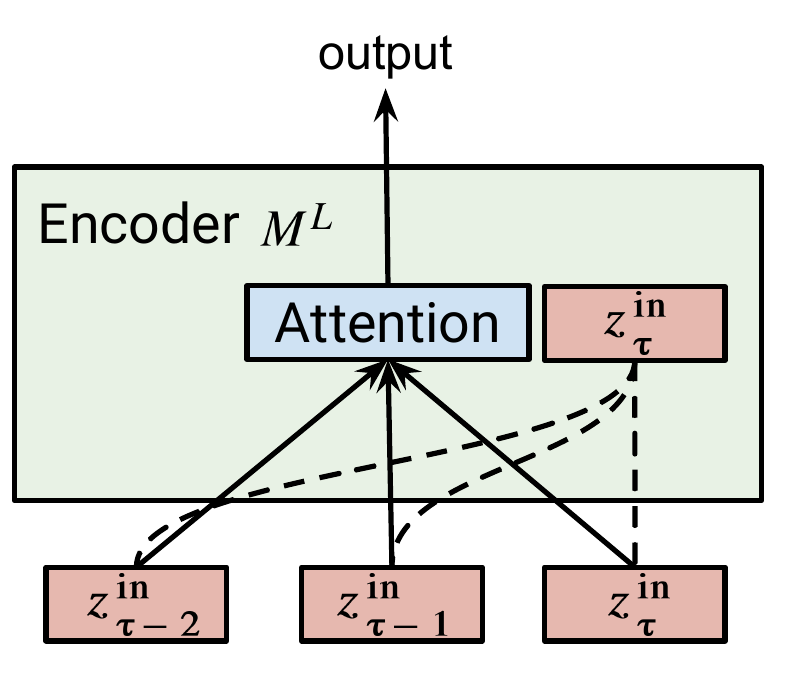}
                \caption{Long-range encoder}
                \label{fig:encoders_ml}
        \end{subfigure}
        % \vspace{-0.2in}
        \caption{The sequence encoders of \emph{M3}.
        The solid lines are used to denote the data flow. The dotted line in (a) means an identity copy whereas in (d) it means the interaction of attention queries and keys.}\label{fig:encoders}
\end{figure*}

Because of its MOE-like structure, our model can adapt to different recommendation scenarios and provide insightful interpretability~(as we shall see in Section~\ref{sec:exper}). 
In many recommendation applications, some features annotate each event and represent the context in which the recommendation query is produced.
Such features are for instance indicative of the page or device on which a user is being served a recommendation.
After obtaining a sequence encoding at step $\tau$~(\emph{i.e.} $SE_\tau$), we fuse it with the annotation's context features~(optional) and project them to the same latent space with another hidden feed-forward layer $F^{\textrm{out}}$:
\begin{equation}
z^{\textrm{out}}_{\tau} = F^{\textrm{out}}([SE_\tau \oplus c^{\textrm{out}}_\tau])
\end{equation}
where $c^{\textrm{out}}$ is a vector encoding contextual information to use after the sequence has been encoded.
Here the $z^{\textrm{out}}_{\tau} \in \mathbb{R}^{1\times d_{\textrm{out}}}$ is what we name \emph{user representation}, it is computed based on the user's history as it has been gathered in logs. 
Finally, a user similarity score $r_v$ is predicted for each item via an inner-product (which can be changed to another similarity scoring function):
\begin{equation}\label{eqn:prediction}
r_v=z^{\textrm{out}}_{\tau} \cdot Q'_v
\end{equation}
where $Q'_v$ is a vector representing the item.
For a given user, item similarity scores are then normalized by a softmax layer which yields a recommendation distribution over the item vocabulary.
After training M3, the recommendations for a user at step $\tau$ are served by sorting the similarity scores $r_v$ obtained for all $v \in \mathcal{V}$ and retrieving the items associated with the highest scores.

\subsection{Three Different Range Encoders}

\textbf{Item Co-occurrence as a Tiny-range Encoder}
The Tiny-range encoder $M^T$ only focuses on the user's last event $e_\tau$, ignoring all previous events. 
In other words, given the processed inputs from past events $Z^{\textrm{in}}_\tau$, this encoder will only consider $z^{\textrm{in}}_\tau$. 
As in factorizing Markov chain~(FMC) models~\cite{rendle2010factorizing}, $M^T$ makes predictions based on item range-1 co-occurrence within observed sequences.
For example, if most of users buy iPhone cases after purchasing an iPhone, then $M^T$ should learn this item-to-item co-occurrence pattern.
As shown in Figure~\ref{fig:encoders_mt}, we compute $M^T$'s output as:
\begin{equation}
M^T(Z^{\textrm{in}}_\tau) = \phi(z^{\textrm{in}}_\tau),
\text{ where }\phi(x) =  \begin{cases}
            xW^{(T)} + b^{(T)},& \text{if } d_{\textrm{in}} \ne d_{\textrm{enc}},\\
            x,              & \text{otherwise}.
        \end{cases}
\end{equation}
That is, when the dimensionality of processed input and encoder output are the same, 
the tiny-range encoder performs a role of residual for the other encoders in mixture. 
%the tiny-range encoder performs a role similar to that of a residual connection. 
If $d_{\textrm{in}} \neq d_{\textrm{enc}}$ , it is possible to down-sample~(if $d_{\textrm{in}}>d_{\textrm{enc}}$) or up-sample (if $d_{\textrm{in}}<d_{\textrm{enc}}$) from $z^{\textrm{in}}_\tau$ by learned parameters $W^{(T)} \in \mathbb{R}^{d_{\textrm{in}}\times d_{\textrm{enc}}}$ and $b^{(T)} \in \mathbb{R}^{d_{\textrm{enc}}}$.

In summary, the tiny-range encoder $M^{T}$ can only focus on the last event by construction, meaning \emph{it has a temporal range of 1 by design}. 
If we only use the output of $M^{T}$ to make predictions, we obtain recommendations results based on item co-occurrence.

\textbf{RNN/CNN as Short-range Encoder}
As discussed in Section~\ref{sec:pre}, the recent behavior of a user has substantial predictive power on current and future interactions.
Therefore, to leverage the corresponding signals entailed in observations, we consider instantiating a short-range sequence encoder that puts more emphasis on recent past events.
Given the processed input from past events $Z^{\textrm{in}}_\tau$, this encoder, represented as $M^S$, focuses by design on a recent subset of logged events. 
Based on our quantitative data exploration, we believe it is suitable for $M^S$ to be highly sensitive to sequence order.
For instance, we expect this encoder to capture the purchasing pattern iPhone $\rightarrow$ iPhone case $\rightarrow$ iPhone charger if it appears frequently in user sequences.
As a result, we believe the Recurrent Neural Network~(RNN~\cite{mikolov2010recurrent}) and the Temporal Convolutional Network~(\cite{van2016wavenet, bai2018empirical, yuan2019simple}) are fitting potential architectural choices.
Such neural architectures have shown superb performances when modeling high-order causalities.
Beyond accuracy, these two encoders are also order sensitive, unlike early sequence modeling method~(\emph{i.e.} Bag-of-Word~\cite{jurafsky2014speech}). 
As a result we develop two interchangeable variants of M3: \emph{M3R} and \emph{M3C} using an RNN and a CNN respectively.

To further describe each of these options, let us introduce our RNN encoder $M^S_{\textrm{RNN}}$. As shown in Figure~\ref{fig:encoders_rnn} we obtain the output of $M^{S}_{\textrm{RNN}}$ by first computing the \emph{hidden state} of RNN at step $\tau$:
\begin{equation}
h_\tau = \text{RNN}(z^{\textrm{in}}_\tau, h_{\tau-1}),
\end{equation}
where $\text{RNN}(\cdot)$ is a \emph{recurrent cell} that updates the hidden state at each step based on the previous hidden state $h_{\tau-1} \in \mathbb{R}^{1 \times d_{\textrm{in}}}$ and the current RNN input $z^{\textrm{in}}_\tau$.
Several choices such as Gated Recurrent Unit~(GRU)~\cite{cho2014learning} or Long Short Term Memory~(LSTM)~\cite{hochreiter1997long} can be used. 
The output is then computed as follows:
\begin{equation}
M^S_{\textrm{RNN}}(Z^{\textrm{in}}_\tau) = h_\tau W^{(R)}, \text{ where } W^{(R)} \in \mathbb{R}^{d_{\textrm{in}} \times d_{\textrm{enc}}}
\end{equation}
where $W^{(R)}$
maps the hidden state to the encoder output space.
We design our CNN encoder $M^S_{\textrm{CNN}}$ as a Temporal Convolutional Networks which has provided state-of-art sequential modeling performance~\cite{van2016wavenet,gehring2017convolutional,bai2018empirical}.
As shown in Figure~\ref{fig:encoders_cnn}, this encoder consists of several stacked layers.
Each layer computes
\begin{equation}
h^{(1)}_\tau = \text{Conv}(Z^{\textrm{in}}_\tau), \dots,
h^{(k)}_\tau = \text{Conv}(h^{(k-1)}_\tau),
\end{equation}
where $k$ indicates the layer number. 
The $\text{Conv}(\cdot)$ is a 1-D convolutional operator~(combined with non-linear activations, see \cite{bai2018empirical} for more details), which contains $d_{\textrm{enc}}$ convolutional filters and operates on the convolutional inputs.
With $K$ layers in our CNN encoder, the final output will be:
\begin{equation}
M^S_{\textrm{CNN}}(Z^{\textrm{in}}_\tau) = h^{(K)}_\tau.
\end{equation}

As highly valuable signals exist in the short-range part of user sequence, we propose two types of encoders to capture them.
Our model can be instantiated in its first variant, \emph{M3R}, if we use RNN encoder or \emph{M3C} if a CNN is employed.
Here M3C and M3R are totally interchangeable with each other and they show comparable results in our experiments~(see Section~\ref{sec:expr_overall}).
We believe such flexibility will help practitioners adapt their model to the hardware they intend to use, \emph{i.e.} typically using GPU for faster CNN training or CPU for which RNNs are better suited.
In terms of temporal range, the CNN only considers a limited finite window of inputs when producing any output. The RNN, although it does not have a finite receptive field, is hampered by difficulties when learning to leverage events located further back into the past (to leverage an event located $L$ observations ago the RNN needs $L-1$ steps).
Regardless of the choice of a CNN or an RNN, our short-range encoder $M^S$ \emph{has a temporal range greater than 1, although it is challenging for this sub-model to capture signals too far away from current step}.
This second encoder is specifically designed to capture sequence patterns that concern recent events.

\begin{table*}[t!]
\center
\caption{A summary of relationships and differences between sequence encoders in \emph{M3}}\label{tb:summary}
\begin{tabular}{lccccc}
\toprule
& \textbf{Base model} & \textbf{Temporal range} & \textbf{Model size} & \textbf{Sensitive to order} & \textbf{Robustness}\\
\midrule
$M^T$ & Item Co-occurrence & 1 & small~(or 0) & very high & no\\
\midrule
$M^S_{\textrm{RNN}}$ & Recurrent Neural Nets & unknown & large & high & no\\
\midrule
$M^S_{\textrm{CNN}}$ & Temporal Convolution Nets & limited & large & high & no\\
\midrule
$M^L$ & Attention Model & unlimited & small~(or 0) & no & high\\
\bottomrule
\end{tabular}
\end{table*}

\textbf{Attention Model as Long-range Encoder}
% Before describing the long-range encoder we setup, let us remind readers of several distinct properties concerning long-range temporal patterns in user sequences.
The choice of an attention model is also influenced by our preliminary quantitative analysis.
As discussed in Section~\ref{sec:pre}, as the temporal distance grows larger, the uncertainties affecting the influence of item consumption on future events get larger as well.
Moreover, as opposed to the recent part of a given user's interaction sequence, relative position does not matter as much when it comes to capturing the influence of temporally distant events. 
% In particular, an event located 100 steps back or 110 steps back may have the same likelihood of influencing the user's current behavior.
As we take these properties into account, we choose to employ \emph{Attention Model}~\cite{bahdanau2014neural, vaswani2017attention} as our long-range sequence encoder. 
Usually, an attention model consists of three parts: attention \emph{queries}, attention \emph{keys} and attention \emph{values}.
One can simply regard an attention model as weighted-sum over attention values with weights resulting from the interaction between attention queries and attention keys. 
In our setting, we use (1) the last event's processed input $z^{\textrm{in}}_\tau$ as attention queries, (2) all past events' processed inputs $Z^{\textrm{in}}_\tau$ as keys and values and (3) scaled dot-product~\cite{vaswani2017attention} as the similarity metric in the attention softmax. 
For instance, if a user last purchased a pair of shoes,
the attention mechanism will focus on footwear related previous purchases.

So that all encoders have the same output dimensionality, we need to transform\footnote{It is unnecessary if $d_{\textrm{in}}$ is same as $d_{\textrm{enc}}$.} our processed input first as follows:
\begin{equation}
\tilde{Z}^{\textrm{in}}_\tau = Z^{\textrm{in}}_\tau W^{(A)},
\end{equation}
where $W^{(A)} \in \mathbb{R}^{d_{\textrm{in}} \times d_{\textrm{enc}}}$ is a learned matrix of parameters. 
% [CAMERA-READY] Add this block according to Review3's suggestion on notations.
Then for each position $i \in [1,\tau]$,
we obtain its raw attention weights, with respect to the processed input $\tilde{z}^{\textrm{in}}_i$, as follows: 
\begin{equation}
\omega_{\tau, i} = \frac{\tilde{z}^{\textrm{in}}_\tau \cdot \tilde{z}^{\textrm{in}}_i}{\sqrt{d_{\textrm{enc}}}},
\end{equation}
where $\omega_{\tau,i}$ is the raw weight at position $i$.
Similarly, we compute the raw attention weights $\boldsymbol{\omega}_\tau \in \mathbb{R}^{1\times \tau}$ for all positions $\boldsymbol{\omega}_\tau=\{\omega_i\}_{i=1..\tau}$ and normalize them with a $\text{softmax}(\cdot)$ function. 
Finally, we acquire the output of our long-range encoder as follows:
\begin{equation}
M^L(Z^{\textrm{in}}_\tau) = \text{softmax}(\boldsymbol{\omega}_\tau) Z^{\textrm{in}}_\tau.
\end{equation}

Our long-range encoder borrows several advantages from the attention model. 
First, it is not limited by a certain finite temporal range.
That is, \emph{it has an unlimited temporal range and can `attend' to anywhere in user's sequence with O(1) steps}. 
Second,because it computes its outputs as a weighted sum of inputs, the attention-based encoder is not as sensitive to sequential order as an RNN or a CNN as each event from the past has an equal chance of influencing the prediction.
Third, the attention model is robust to noisy inputs due to its normalized attention weights and weighted-sum aggregation. 

\textbf{Gating Network}
Borrowing the idea from from Mixture-of-Experts model~\cite{jacobs1991adaptive, ma2018mtr}, we build a gating network to aggregate our encoders' results. The gate is also helpful to better understand our model~(see Section~\ref{sec:exper}). 
To produce a simpler gating network, we use a feed-forward layer $F^{g}$ on the gating network's inputs:
\begin{equation}
G_\tau = [G^T_\tau, G^S_\tau, G^L_\tau] = \text{sigmoid}(F^{g}(G^{\textrm{in}}_\tau)),
\end{equation}
where $G^{\textrm{in}}_\tau$ is the input we feed into our gating network. 
We will discuss how the model performs overall with different choices of gate inputs in Section~\ref{sec:expr_gate}.
The resulting $G_\tau \in \mathbb{R}^3$ contains the gate value modulating each encoder. 
More importantly, an element-wise sigmoid function is applied to the gate values which allows encoders to `corporate' with each other~\cite{belletti2018factorized}.
Note that a few previous works~\cite{ma2018mtr, jordan1994hierarchical, shazeer2017outrageously} also normalize the gate values, but we found this choice led to the degeneracy of our mixture model as it would learn to only use $M^S$ which in turn hampers model performance.

\textbf{Summary}
% We summarize the relationships and differences between our sequence encoders in  
M3 is able to address limitations of pre-existing models as shown in Table~\ref{tb:summary}: 
(1) M3 has a mixture of three encoders with different temporal ranges which can capture sequential patterns located anywhere in user sequences. 
(2) Instead of learning a set of latent factor $P_u$ for each user, M3 represents the long-term user preferences by using a long-range sequence encoder that provides a representation of the entire history of a user. 
Furthermore, M3 is efficient in both model size and computational cost. 
In particular M3 does not introduce any extra parameters under certain settings (\emph{i.e.} $d_{\textrm{in}}=d_{\textrm{enc}}$), and the computation of $M^T$ and $M^L$ are very efficient when using specialized hardware such as a GPU. 
With its simple gate design, M3 also provides good interpretability and adaptability.

\begin{itemize}
    \item \textbf{Effectiveness.} Given our analysis on user sequences, we assume M3 to be effective. As compared to past works, M3 is capable to capture signals from the whole sequence, it also satisfies the properties we found in different parts of sequence. Moreover, our three encoders constitute a diverse set of sequential encoder and, if well-trained, can model user sequence in a multi-scale manner, which is a key to success in past literature~\cite{van2016wavenet,yu2015multi}. 
    %Our empirical evaluation results in Section~\ref{sec:exper} verify our assumption.
    \item \textbf{Efficiency.} In terms of model size, M3 is efficient. As compared to existing works which use short-range encoder only, though uses two other encoders, our M3 model doesn't introduce any extra parameters~(if $d_{\textrm{in}}=d_{\textrm{enc}}$). In terms of computational efficiency, our M3 is good as well, as both $M^T$ and $M^L$ are nothing other than matrix multiplication, which is cheap when computed with optimized hardwares like Graphics Processing Unit~(GPU).
    \item \textbf{Interpretability.} Model's interpretability is critical for diagnosing purpose. As we shall see later, with the gate network, we are able to visualize our network transparently by observing the gate values. 
    \item \textbf{Adaptability.} One issue in production recommender system is modeling users for different recommendation scenarios, as people may behave very differently. Two typical scenarios are \emph{HomePage} recommendation and product \emph{DetailPage} recommendation. However, as we shall introduce in later section, M3 is able to adapt to these scenarios if we use the scenario information as our gate input.
\end{itemize}
\section{Experimental Studies}\label{sec:exper}
In this section, we study the two variants of M3 against several baseline state-of-the-art methods on both a publicly available dataset and our large-scale Youtube dataset.

\textbf{Datasets} 
We use MovieLens 20M\footnote{https://grouplens.org/datasets/movielens/20m/}, which is a publicly available dataset, along with a large-scale anonymized dataset from YouTube 
to which we have access because we are employees of Google working on improving YouTube as a platform.
The dataset is private, anonymized and accessible only internally by few employees whose work is directly related to Youtube.

\subsection{Experiments on MovieLens Dataset}
As in previous works~\cite{tang2018caser,HeMcA16b}, we process the MovieLens data by first converting numeric ratings to 1 values, turning them into implicit logged item consumption feedback. We remove the items with less than $20$ ratings. 
Such items, because of how little user feedback is available for them, represent another research challenge --- cold start --- which is outside the scope of the present paper.

To focus on long user sequences, we filtered out users who had a sequence length of less than $\delta_{\text{min}}=20$ item consumed, while we didn't filter items specifically.
The maximum sequence length in the dataset being $7450$, we follow the method proposed in~\cite{HeMcA16b,tang2018caser} and employ a sliding window of length $\delta_{\text{win}}=300$ to generate similarly long sequences of user/item interactions in which we aim to capture long range dependent patterns. Some statistics can be found in the first row of Table~\ref{tab:variants}.

We do not use contextual annotations for the MovieLens data.

\textbf{Evaluation protocol}
We split the dataset into training and test set by randomly choosing 80\% of users for training and the remaining 20\% for validation (10\%) and testing (10\%). 
As with the training data, a sliding window is used on the validation and test sets to generate sequences.
We measure the mean average precision (mAP)
as an indicator for models' performances~\cite{beutel2017beyond, tang2018ranking}. We only focus on the
top positions of our predictions, so we choose to use mAP@n with
$n \in \left\{5, 10, 20\right\}$.
There is only one target per instance here and therefore the mAP@n is expected to increase with $n$ which is consistent with~\cite{belletti2018factorized} but differs from~\cite{tang2018caser}.

\textbf{Details on model architectures}
We keep architectural parameters consistent across all experiments on MovieLens. In particular, we use identical representation dimensions: 
$
    d_{\textrm{in}}
        =
    d_{\textrm{enc}}
        =
    d_{\textrm{out}}
        =
    32.
$
Such a choice decreases the number of free parameters as the sub-models $M^T$ and $M^L$ will not have learned parameters.
A GRU cell is employed for the RNN while 2 stacked temporal convolution layers~\cite{bai2018empirical} of width 5 are used in the CNN.
A ReLU activation function is employed in the feed-forward layers $F^{\textrm{in}}$ and $F^{\textrm{out}}$.
Item embeddings of dimension $64$ are learned with different weights on the input side~(\emph{i.e.,} $Q$ in Eq.~\ref{eqn:input}) and output side~(\emph{i.e.,} $Q'$ in Eq.~\ref{eqn:prediction}).
Although previous work~\cite{HeMcA16b} has constrained such embeddings to be identical on the input and output side of the model, we found that increasing the number of degrees of freedom led to better results.

\textbf{Baselines}
We compare our two variants, \emph{i.e.,} M3R and M3C, with the following baselines:
\begin{itemize}
    \item \textbf{FMC}: The Factorizing model for the first-order Markov chain~(FMC)~\cite{rendle2010factorizing} is a simple but strong baseline in sequential recommendation task~\cite{tang2018caser, beutel2018latent, smirnova2017context}. As discussed in Section~\ref{sec:intro}, we do not want to use explicit user representations. Therefore, we do not compare the personalized version of this model~(FPMC).
    \item \textbf{DeepBoW}: The Deep Bag-of-word model represent user by averaging item embeddings from all past events. The model then makes predictions through a feed-forward layer. In our experiments, we use a single hidden layer with size of 32 and ReLU as activation function.
    \item \textbf{GRU4Rec}: Originally presented in~\cite{hidasi2015session}, this method uses a GRU RNN over user sequences and is a state-of-the-art model for sequential recommendation with anonymized data.
    \item \textbf{Caser}: The Convolutional Sequence Embeddings model~\cite{tang2018caser} applying horizontal and vertical convolutional filters over the embedding matrix and achieves state-of-the-art sequential recommendation performance. We try $\left\{2, 4, 8\right\}$ vertical filters and $\left\{16, 32, 64\right\}$ horizontal filters of size $(3, 5, 7)$.
    In order to focus on the sequential encoding task, we discard the user embedding and only use the sequence embedding of this model to make predictions.
\end{itemize}

In the models above, due to the large number of items in input and output dictionaries, the learned embeddings comprise most of the free parameters.
Therefore, having set the embedding dimension to $64$ in all the baselines as well as in M3R and M3C, 
we consider models with similar numbers of learned parameters.
The other hyperparameters mentioned above are
tuned by looking at the mAP@20 on validation set. 
% [CAMERA-READY] Add training time description according to Review2's suggestion.
The training time of M3R/M3C is comparable with others and can be further improved with techniques like model compression~\cite{tang2018ranking}, quantization~\cite{hubara2017quantized}, etc.

\subsubsection{Overall Performances}
We report each model's performance
in Table~\ref{tab:ML20Mresults}. 
Each metric is averaged across all user sequences in test set. The best performer is highlighted in bold face. 
The results show that both M3C and M3R outperform other baselines by a large margin. 
Among the baselines, GRU4Rec
achieves the best performance and DeepBoW worst one,
suggesting the sequence order plays a very important predictive role. 
FMC performs surprisingly well, suggesting
we could get considerable results with a simple model only taking the last event into account. 
The poor results of Caser may be caused by its design which relies on vertical filters of fixed size. Caser performs better in the next subsection which considers sequences whose lengths vary less within the training data.

\begin{table}[h!]
    \centering
    \begin{tabular}{cccc}
    \toprule
         &  \textbf{mAP@5} & \textbf{mAP@10} & \textbf{mAP@20} \\
    \midrule
        \textbf{FMC} & $0.0256$ & $0.0291$ & $ 0.0317$ \\
        \textbf{DeepBoW} & $0.0065$ & $0.0079$ & $0.0093$ \\
        \textbf{GRU4Rec} & $0.0256$ & $0.0304$ & $0.0343$ \\
        \textbf{Caser} & $0.0225$ & $0.0269$ & $0.0304$ \\
    \midrule
        \textbf{M3C} & $0.0295$ & $0.0342$ & $0.0379$ \\
        \textbf{M3R} & $\textbf{0.0315}$ & $\textbf{0.0367}$ & $\textbf{0.0421}$ \\
    \midrule
        Improv. & $+23.4\%$ & $+20.7\%$ & $+22.7\%$ \\
    \bottomrule
    \end{tabular}
    \caption{Performance comparison on MovieLens 20M. M3C and M3R outperform the baselines significantly.}
    \label{tab:ML20Mresults}
\end{table}

\subsubsection{Investigating the influence of sequence length through variants of MovieLens}
The previous results have shown strong performance gains achieved by the models we intruced: M3C and M3R.
We now investigate the origin of such improvements.
The design of these models was inspired by an attempt to capture sequential patterns with different characteristic temporal extents.
To check whether the models we introduced achieve this aim we construct multiple variants of MovieLens with different sequence lengths.

We vary the sequence length by having a maximum cutoff threshold $\delta_{\textrm{max}}$ which complements the minimal sequence length threshold $\delta_{\textrm{min}}$.
A sequence with more than $\delta_{\textrm{max}}$ only has its latest $\delta_{\textrm{max}}$ observations remained.
We vary the values of $\delta_{\textrm{min}}$, $\delta_{\textrm{max}}$ and the sequence generation window size.
Table~\ref{tab:variants} summarizes the properties of the four variants of the MovieLens dataset we construct. It is noteworthy that such settings make Caser perform better as the sequence length is more consistent within each dataset variant.

\begin{table*}[h!]
    \centering
    \begin{tabular}{lcccccc}
    \toprule
    & \textbf{Min. Length} & \textbf{Max. Length} & \textbf{Window Size} & \textbf{Avg. Length} & \textbf{\#Sequences} & \textbf{\#Items} \\
    \midrule
    \textbf{ML20M} & 20 & $\infty$ & 300 & 144.1 & 138.4K & 13.1K \\
    \midrule
    \textbf{ML20M-S} & 20 & 50 & 20 & 42.8 & 138.4K & 13.1K \\
    \midrule
    \textbf{ML20M-M} & 50 & 150 & 50 & 113.6 & 85.2K & 13.1K \\
    \midrule
    \textbf{ML20M-L} & 150 & 300 & 150 & 250.7 & 35.8K & 12.9K \\
    \midrule
    \textbf{ML20M-XL} & 300 & $\infty$ & 300 & 605.5 & 16.3K & 12.5K \\
    \bottomrule
    \end{tabular}
    \caption{Statistics of the variants of the MovieLens dataset.}
    \vspace{-0.3in}
    \label{tab:variants}
\end{table*}

\begin{figure*}
    \centering
    \includegraphics[width=\textwidth]{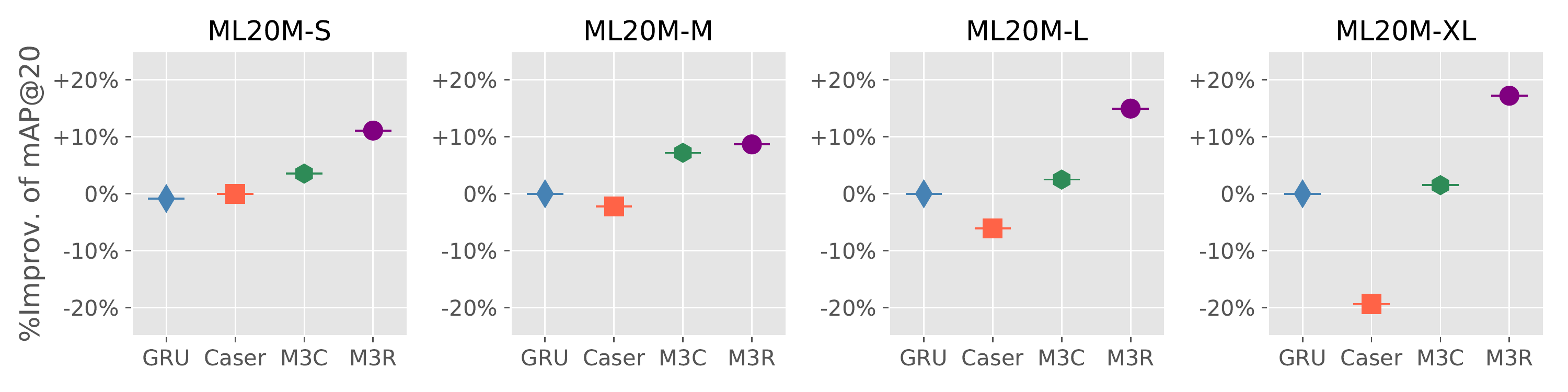}
    \caption{
    Uplifts with respect to the best baselines on four MovieLens variants. The improvement percentage of each model is computed by its relative mAP@20 gain against the best baseline. 
    For all variants, M3R significantly outperforms the two baselines we consider according to a one-tail paired t-test at level $0.01$, while M3C outperforms the other two significantly only on ML20M-M. Note that the standard error of all uplifts gets higher as we use a MovieLens variant with longer sequences. The standard error reaches $2.3\%$ on ML20M-XL.}
    \label{fig:variants}
\end{figure*}

GRU4Rec and Caser outperform the other baselines in the present setting and therefore we only report their performance.
Figure~\ref{fig:variants} shows the
improvements of M3C and M3R over the best baselines on four MovieLens variants.
The improvement of each model is computed
by its mAP@20 against the best baseline. 
In most cases, M3C and
M3R can outperform the highest performing baseline.
Specifically, on ML20M-S and ML20M-M, Caser performs similarly
to GRU4Rec while both M3C and M3R have good performance. This is probably due to the contribution of the tiny-range encoder.

\subsection{Anonymized YouTube dataset}
For the YouTube dataset, we filtered out users whose logged sequence length was less than 150 ($\delta_{\textrm{min}}=150$) and keep each user's last 300 events ($\delta_{\textrm{max}}=300$) in their item consumption sequence.
In the following experiments, we exploit contextual annotations such as user device~(\emph{e.g.,} from web browser or mobile App), time-based features~(\emph{e.g.,} dwelling time), etc.
User sequences are all anonymized and precautions have been taken to guarantee that users cannot be re-identified. In particular, only public videos with enough views have been retained. 

Neural recommender systems attempt at foreseeing the interest of users under extreme constraints of latency and scale. We define the task as predicting the next item the user will consume given a recorded history of items already consumed. Such a problem setting is indeed common in collaborative filtering~\cite{sarwar2001item,linden2003amazon} recommendations. 
We present here results obtained on a dataset where only about 2 million items are present that correspond to most popular items. 
While the user history can span over months, only watches from the last 7 days are used for labels in training and watches in the last 2 days are used for testing. The train/test split is $90/10\%$.
The test set does not overlap with the train set and corresponds to the last temporal slice of the dataset.
In all, we have more than 200 million training sequences and more than 1 million test sequences, and with overall average sequence length approximately being 200.

The neural network predicts, for a sample of negatives, the probability that they are chosen and classically a negative sampling loss is employed in order to leverage observations belonging to a very large vocabulary~\cite{jean2014using}.
The loss being minimized is
$$
\sum_{l \in \text{Labels}} w_l \times \text{CrossEntropy}(\text{SampledSoftmax}(\xi(t+1)))
$$
where the SampledSoftmax~\cite{jean2014using} uses $20000$ randomly sampled negatives and $w_l$ is the weight of each label. 

\textbf{Evaluation metrics} To test the models' performances, we measure the mean average precision~(mAP) as in~\cite{tang2018caser, beutel2018latent}. We only focus on the top positions of our predictions, so we choose to use mAP@$n$ with $n \in \{5, 10, 20\}$.
The mAP is computed with the entire dictionary of candidate items as opposed to the training loss which samples negatives.
There is only one target per instance here and therefore the mAP@n is expected to increase with n which is consistent with~\cite{belletti2018factorized} but differs from~\cite{tang2018caser}.
% In this setting the mAP@n is not divided by $n$ which is why it increases as $n$ increases.

\begin{table}[t]
\caption{Performance comparison on the anonymized YouTube dataset. M3C and M3R outperform the baselines significantly.}
\centering
\label{tb:ytperform}
\begin{tabular}{cccc}
\toprule
& \textbf{mAP@5} & \textbf{mAP@10} & \textbf{mAP@20} \\
\midrule
\textbf{Context-FMC} & $0.1103$ & $0.119$ & $0.1240$ \\
\textbf{DeepYouTube} & $0.1295$ & $0.1399$ & $0.1455$ \\
\textbf{Context-GRU} & $0.1319$ & $0.1438$ & $0.1503$ \\
\midrule
\textbf{M3C} & $0.1469$ & $0.1591$ & $0.1654$ \\
\textbf{M3R} & $\textbf{0.1541}$ & $\textbf{01670}$ & $\textbf{0.1743}$ \\
\midrule
Improv. & $+16.8\%$ & $+16.1\%$ & $+16.0\%$ \\
\bottomrule
\end{tabular}
\end{table}

\textbf{Baselines}
In order to make fair comparisons with all previous baselines,
we used their contextual counterparts if they are proposed or
compared in literature.
\begin{itemize}
    \item \textbf{Context-FMC}: The Context-FMC condition the last event's
embedding on last event's context features by concatenating
them and having a feed-forward layer over them.
    \item \textbf{DeepYouTube}: Proposed by~\cite{paul2016deep}, the DeepYoutube model is a state-of-the-art neural model for recommendation. It concatenates: (1) item embedding from users' last event, (2) item embeddings averaged by all past events and (3) context features. The model then makes predictions through a feedforward layer composed of several ReLU layers.
    \item \textbf{Context-GRU}: We used the contextual version of GRU proposed in~\cite{smirnova2017context}. Among the three conditioning paradigms on context, we used the concatenation as it gives us better performances.
\end{itemize}
All models are implemented by TensorFlow~\cite{abadi2016tensorflow} and by Adagrad~\cite{duchi2011adaptive} over a parameter server~\cite{li2014scaling} with many workers. 

\textbf{Model details}
In the following experiments, we keep the dimensions of processed input $d_{\textrm{in}}$ and encoder outputs $d_{\textrm{enc}}$ identical for all experiments conducted on the same dataset.
Once more, we also want to share some of our architectural parameters so that they are consistent across the two datasets.
Again, by doing this, we make the parametrization of our models more parsimonious, because the sub-models $M^T$ and $M^L$ will be parameter-free.
For the RNN cell, we use a GRU on both datasets for its effectiveness as well as efficiency. 
For the CNN version, we stacked $3$ layers of temporal convolution~\cite{bai2018empirical}, with no dilation and width of $5$.
For the feed-forward layers $F^{\textrm{in}}$ and $F^{\textrm{out}}$, we used ReLU as their activation functions, whereas they contains different number of sub-layers.
For item embeddings on the input side~(\emph{i.e.,} $Q$ in Eq.~\ref{eqn:input}) and on the output side~(\emph{i.e.,} $Q'$ in Eq.~\ref{eqn:prediction}), we learn them separately which improves all results.

\subsubsection{Overall Results}\label{sec:expr_overall}
We report each model's performance on the private dataset in Table~\ref{tb:ytperform}. 
The best performer is highlighted in bold face. 
As can be seen from this table, on our anonymized YouTube dataset, the Context-FMC performs worse followed by DeepYoutube while Context-GRU performs best among all baselines. 
The DeepYouTube and Context-GRU perform better than Context-FMC possibly because they have longer
temporal range, which again shows that the temporal range matters significantly in long user sequences.
One can therefore improve the performance of a sequential recommender if the model is able to leverage distant~(long-range dependent) informantion in user sequences.

On both datasets, we observed our proposed two model variants M3R and M3C significantly outperform all other baselines.
Within these two variants, the M3R preforms marginally better than the M3C, and it improves upon the best baselines by a large margin (more than 20\% on MovieLens data and 16.0\% on YouTube data).

\begin{table}[t!]
\center
\caption{mAP@20 vs. different components of M3R on both datasets, where T,S,L stands for $M^T$, $M^S$ and $M^L$ respectively. }\label{tb:ablation}
\setlength{\tabcolsep}{4pt}
\begin{tabular}{lcc}
\toprule
 & \textbf{MovieLens 20M} & \textbf{YouTube Dataset}\\
\midrule
M3R-T & 0.0269 & 0.1406\\
\midrule
M3R-S & 0.0363 & 0.1673\\
\midrule
M3R-L & 0.0266 & 0.1359\\
\midrule
M3R-TS & 0.0412 & 0.1700\\
\midrule
M3R-TL & 0.0293 & 0.1485\\
\midrule
M3R-SL & 0.0403 & 0.1702\\
\midrule
M3R-TSL & 0.0421 & 0.1743\\
\bottomrule
\end{tabular}
\end{table}

\subsection{Ablation Study of Mixture of Models}\label{sec:expr_ablation}
To demonstrate how each encoder contributes to the overall performance, we now present an ablation test on our M3R model (results from M3C are similar) on our proprietary data.
We use {T, S, L} to denote $M^T$, $M^S$ and $M^L$ respectively. 
The results are described in Table~\ref{tb:ablation}. 
When we only enable single encoder for M3R, the best performer is M3R-T on MovieLens data and M3R-S on the YouTube data. 
This result is consistent with the results in Section~\ref{sec:expr_overall}. 
With more encoders involved in M3R the model performs better.
In particular, when all encoders are incorporated, our M3R-TSL performs best on both datasets, indicating all three encoders matter for performance.

\subsection{Role of Gating Network}\label{sec:expr_gate}
We now begin to study our gating network in order to answer the following questions: 
(1) Is the gating network beneficial to the overall model performance? 
(2) How do different gating network inputs influence the model performance? and 
(3) How can the gating network make our model more adaptable and interpretable?

\begin{figure}[t!]  %figure divergent
    \includegraphics[width=0.45\textwidth]{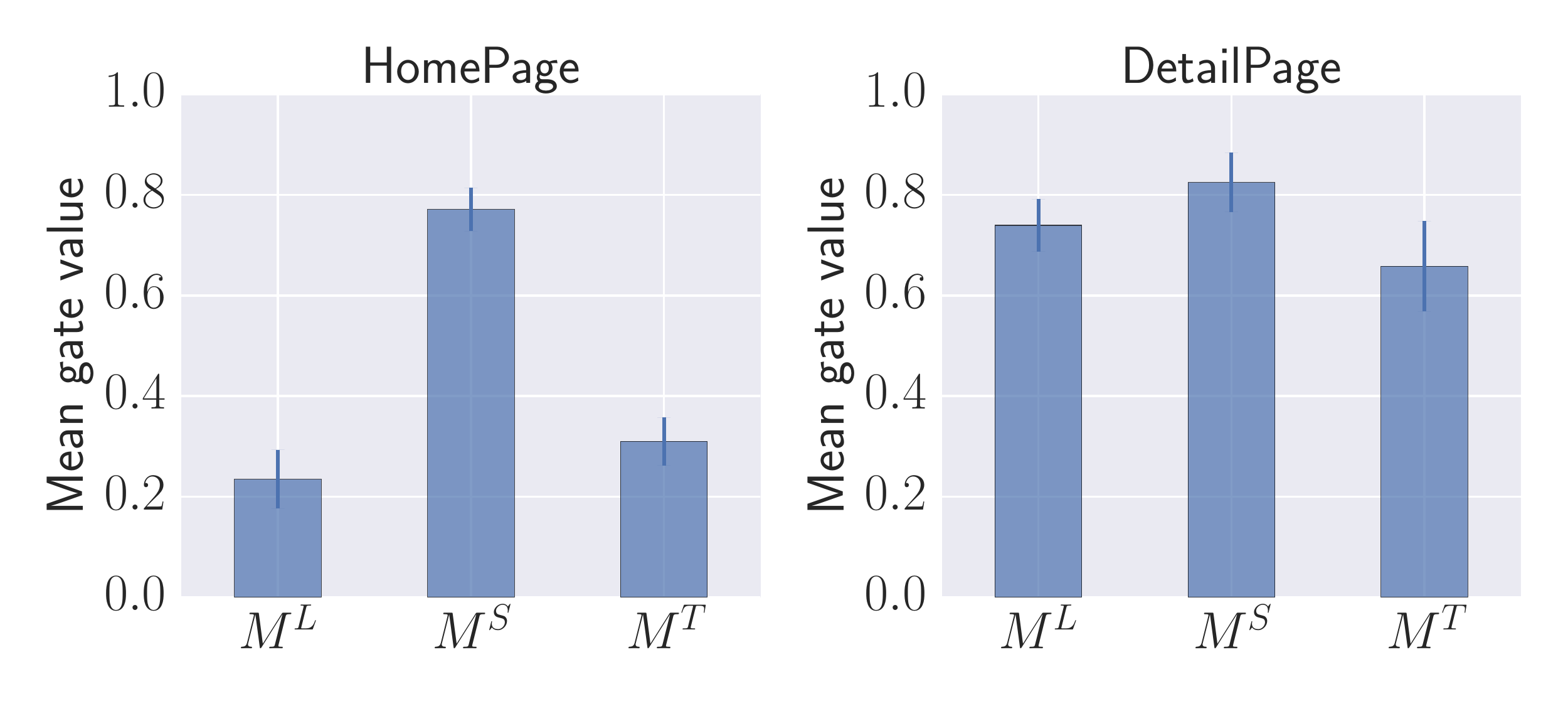}
    \vspace{-0.5cm}
    \caption{Average gate values of M3R in different scenarios. The model learns to use different combination of encoders in different recommendation scenarios.}\label{fig:gate_visualize}
\end{figure}

\textbf{Fixed gates versus learned gates:}
First of all, we examine the impact of our gating network by comparing it with a set of fixed gate values.
More precisely, we fixed the gate values to be all equal to $1.0$ during the model training: $G_\tau = \boldsymbol{1}$, here $\boldsymbol{1} \in \mathbb{R}^{3}$ is a vector. 
The first row of Table~\ref{tb:gate} shows the result of this fixed-gate model.
We found that the fixed models are weaker than the best performing version of M3R~(\emph{i.e.,} mAP@20 of 0.1743) and M3C~(\emph{i.e.,} mAP@20 of 0.1654).
This reveals that the gating network consistently improves M3-based models' performances. 

\textbf{Influence of different gate inputs:}
In this paragraph we investigate the potential choices of inputs for the gating network, and how they result in different performance scores. 
In the existing Mixture-of-Experts~(MOE) literature, the input for the gating network $G^{\textrm{in}}_\tau$ can be categorized into Contextual-switch and Bottom-switch.
The Contextual-switch, used in~\cite{belletti2018factorized}, uses context information as gate input:
\begin{equation}
G^{\textrm{in}}_\tau=[c^{\textrm{in}}_\tau \oplus c^{\textrm{out}}_\tau],
\end{equation}
where $c^{\textrm{in}}_\tau$ and $c^{\textrm{out}}_\tau$ are context features from input and output side. 
Intuitively, this suggests how context may influence the choices of different encoders.
If no context information is available, we can still use the output of a shared layer operating before the MOE layer~\cite{ma2018mtr, shazeer2017outrageously} as gate input, \emph{i.e.,} Bottom-switch:
\begin{equation}
G^{\textrm{in}}_\tau=z^{\textrm{in}}_\tau.
\end{equation}
The shared layer contains high-level semantic knowledge from the last event, which can also enable gate switching.

On the MovieLens data, we used Bottom-switched gate for all the results above because of the absence of contextual annotations.
On the YouTube dataset, the last two rows from Table~\ref{tb:gate} provide the comparison results between Contextual-switched gate and Bottom-switched gate.
We observe that context information is more useful to the gates than a shared layer. 
In other words, the decision of whether to focus more on recent part~(\emph{i.e.} large gate values for $M^T$ and $M^S$) or on the distant part~(\emph{i.e.} large values for $M^L$) from user sequence is easier to make based on contextual annotations.

\subsection{Discussion on model architecture}
The model architecture we design is based on quantitative findings and has two primary goals: capturing co-existing short-range and long-range behavioral patterns as well as serving recommendations given in different contexts with a single model. 

We know for recommender systems in most applications~(\emph{e.g.} e-commerce like Amazon, streaming services like Netflix) that
recommendations commonly occur in at least two different contexts: either a \emph{HomePage} or a \emph{DetailPage}.
The Homepage is the page shown when users open the website or open the mobile App, while DetailPage is the page shown when users click on a certain item.
User behaviors are different depending on which of these two pages they are browsing.
Users are more likely to be satisfied by a recommendation related to recent events, especially the last event, when they are on a DetailPage.
A straightforward solution to deal with these changing dynamics is to train two different models.

We now demonstrate that with the multi-temporal-range encoders architecture and gating mechanism in M3, we can have a single adaptive end-to-end model that provides good performance in a multi-faceted recommendation problem.
To that end we analyze the behavior of our gating network and show the adaptability of the model as we gain a better understanding of its behavior.

\begin{table}[t!]
\center
\caption{mAP@20 vs. different types of gating network on the two datasets for M3R. `Fixed' indicates we fix gate values to $1.0$, `Contextual-switch' means that we use context features $c^{\textrm{in}}$ and $c^{\textrm{out}}$ as gate input and `Bottom-switch' corresponds to the use of $z^{\textrm{in}}_\tau$ as gate input.}\label{tb:gate}
\setlength{\tabcolsep}{4pt}
\begin{tabular}{lcc}
\toprule
 & \textbf{MovieLens} & \textbf{YouTube}\\
\midrule
Fixed & 0.0413 & 0.1715\\
\midrule
Bottom-switch & 0.0421 & 0.1734\\
\midrule
Contextual-switch & / & 0.1743\\
\bottomrule
\end{tabular}
\end{table}

What we observe in Figure~\ref{fig:gate_visualize} is that when contextual information is available to infer the recommendation scenario, the gating network can effectively automatically decide how to combine the results from different encoders in a dynamic manner to further improve performance.
Figure~\ref{fig:gate_visualize} shows how gate values of M3R change \emph{w.r.t.} across different recommendation scenarios.
It is clear that M3R puts more emphasis on $M^S$ when users are on the HomePage, while it encourages all three encoders involved when users are on DetailPage.
This result shows that the gating network uses different combinations of encoders for different recommendation scenarios.

As a result, we can argue that our architectural design choices do meet the expectations we set in our preliminary analysis.
It is noteworthy that the gating mechanism we added on top of the three sub-models is helpful to improve predictive performance and ease model diagnosis.
We have indeed been able to analyze recommendation patterns seamlessly.
\section{Conclusion}
\label{sec:conclu}

M3 is an effective solution to provide better recommendations based on long user sequences.
M3 is a neural model that avoids most of the limitations faced by pre-existing approaches and is well adapted to cases in which short term and long term temporal dependencies coexist. 
Other than effectiveness, this approach also provides several advantages such as the absence of a need extra parameters and interpretability.
Our experiments on large public dataset as well as a large-scale production dataset suggest that M3 outperforms the state-of-the-art methods by a large margin for sequential recommendation with long user sequences.
One shortcoming of the architecture we propose is that all sub-models are computed at serving time.
As a next step, we plan to train a sparse context dependent gating network to address this shortcoming.

\bibliographystyle{ACM-Reference-Format}
{\small
\bibliography{ref_wsdm19}
}

\end{document}